\documentclass[11pt]{article}

\usepackage[]{acl}

\usepackage{times}
\usepackage{latexsym}
\usepackage{multirow}

\usepackage{graphicx}
\usepackage{amsmath}
\usepackage[most]{tcolorbox}
\usepackage{amsfonts}
\usepackage{amssymb}  
\usepackage{booktabs} 
\usepackage{multirow} 
\usepackage{tabularx}
\usepackage[T1]{fontenc}

\usepackage[utf8]{inputenc}

\usepackage{microtype}

\usepackage{inconsolata}

\usepackage{graphicx}

\usepackage{enumitem}
\usepackage{dblfloatfix}

\title{\textsc{PersonaEdit}: Representative Sample Selection for Personalized Model Editing}

\author{
    You-Mei Huang,\textsuperscript{\rm 1}
    Chung-Chi Chen,\textsuperscript{\rm 2}
    An-Zi Yen\textsuperscript{\rm 1}
    \\
    \textsuperscript{\rm 1} Department of Computer Science, National Yang Ming Chiao Tung University, Taiwan
    \\
    \textsuperscript{\rm 2} Human-Agent Ally Lab (HAA Lab), National Institute of Informatics (NII), Japan
    \\
    \texttt{ymh636.cs13@nycu.edu.tw,}
    \texttt{chen@nii.ac.jp,}
    \texttt{azyen@nycu.edu.tw}
}

\begin{document}
\maketitle
\begin{abstract}
Personalization has attracted growing interest in LLM applications, yet existing retrieval-based approaches depend heavily on retrieval quality and degrade in long-term interactions. 
Model editing, which directly modifies internal model parameters to incorporate new knowledge, has demonstrated effective knowledge modification capabilities in factual knowledge editing tasks and may provide a potential solution for personalization.
However, scaling model editing to personalization is non-trivial. 
Editing large amounts of user data increases computational cost and causes interference among edits, motivating the need for effective sample selection. 
To address this issue, we propose, \textsc{PersonaEdit}, a hidden representation clustering strategy that selects representative editing samples through proportional stratified sampling.
Experiments show that model editing is effective for personalization, and that our selection strategy preserves most of the performance while substantially reducing the number of required editing samples. 
Beyond standalone editing, we find that combining model editing with retrieval-based prompt augmentation further improves personalization, as edited knowledge and retrieved context provide complementary information. 
These results demonstrate the potential of model editing as an efficient and scalable approach for LLM personalization.
\end{abstract}

\section{Introduction}



With the rapid advancement of large language models (LLMs) in dialogue systems and user-interactive applications, there has been growing interest in personalized LLM systems that adapt responses according to individual user preferences and behaviors.
Personalization has shown potential for improving user satisfaction and response quality in personalized response generation tasks \citep{zhang2026personalize, qiu2025measuring}.
Existing approaches commonly rely on context-based methods, including In-context Learning (ICL) and Retrieval-Augmented Generation (RAG) \citep{zhang2026personalize, qiu2025measuring, kim2025few}.
However, these methods suffer from three fundamental limitations: 
(1) strong dependence on retrieval quality and prompt engineering \citep{lewis2020retrieval}; 
(2) inherent constraints imposed by the limited context window \citep{ratner2023parallel};
(3) performance degradation in long-term interaction settings \citep{tan2025prospect}, as the model lacks the ability to truly internalize user preferences.

Recent studies have begun to explore more structured personalization approaches, such as user embeddings, soft prompt tuning and representation editing \citep{doddapaneni2024user,liu2025llms+,zhang2025personalize}.
Among these approaches, model editing provides a promising direction by directly modifying model parameters to incorporate user-specific knowledge and preferences into the model, potentially enabling more persistent personalized behavior without relying entirely on external retrieval during inference \citep{meng2022locating}.
However, compared with factual knowledge editing, the application of model editing to user-level personalization remains relatively less explored.
Moreover, incorporating all user interaction data into the editing process introduces significant computational overhead and scalability challenges \citep{gupta2024model}. 
Therefore, this study investigates the following question: \textit{Can effective personalization be achieved by editing the model using only a small set of representative samples?}

To address this issue, we investigate sample selection strategies for personalization through model editing, including topic-based and hidden representation approaches.
Using OpinionQA \citep{santurkar2023whose}  as a testbed, we analyze how different selection strategies influence the generalization behavior of model editing.
Our analysis shows that human-annotated topics are insufficient to effectively characterize editing generalization, while hidden representation similarity better captures transferable editing behavior for sample selection.
Further experiments demonstrate that the proposed approach improves sample efficiency while preserving personalization capability.

In addition, we observe that personalization performance varies across users and is influenced by both user stance consistency and the discrepancy between model preferences and user preferences.

Our contributions are summarized as follows:

\begin{itemize}[nosep]
    \item \textbf{Applying model editing to personalization:} We demonstrate  that model editing can effectively improve personalization performance, and that incorporating additional user information, such as persona descriptions and few-shot examples, further enhances performance.\footnote{The prompts and code are available at at \url{https://github.com/NYCU-NLP-Lab/PersonaEdit}}
    
    \item \textbf{Hidden representation-based sample selection for model editing:} We propose a hidden representation clustering strategy for selecting representative editing samples. 
    Experimental results show that the approach reduces required editing samples while preserving personalization performance.
    
    \item \textbf{Analysis of user factors affecting model editing effectiveness:} We analyze how user-specific characteristics, including preference consistency and preference gaps between users and models, affect the effectiveness of model editing-based personalization.
\end{itemize}

\section{Related Work}

\subsection{LLM Personalization}

LLM personalization aims to generate responses that better align with individual users’ preferences, viewpoints, or behavioral patterns. 
Early approaches primarily rely on prompt engineering or profile conditioning, incorporating user profiles and historical interaction records into prompts, or leveraging LLMs to iteratively construct personalized prompts to guide response generation \citep{kim2025few}.
These methods are training-free and thus do not require additional model fine-tuning; however, their effectiveness is highly dependent on prompt design quality \citep{sclar2024quantifying} and the model’s in-context reasoning capabilities.

More recent approaches increasingly adopt Retrieval-Augmented Generation (RAG) \citep{qiu2025measuring}, which retrieves user-relevant information during inference and incorporates it into the context window to enhance personalization performance. 
However, retrieval-based approaches still suffer from several limitations, including strong dependence on retrieval quality, constraints imposed by the limited context window \citep{ratner2023parallel}, and performance degradation in long-term interaction settings \citep{tan2025prospect}.

Beyond retrieval-based methods, recent studies have also explored latent personalization approaches, which represent user preferences in latent space through mechanisms such as user embeddings \citep{liu2025llms+}, soft prompts \citep{qiu2025latent}, and representation editing \citep{zhang2025personalize}. 

However, these methods still rely on latent conditioning or representation-level intervention during inference, requiring personalized information to be provided or maintained externally rather than incorporated into the model itself. 
This motivates the exploration of approaches that directly encode personalized knowledge into model parameters, such as model editing.

\subsection{Model Editing}
Model editing aims to inject new knowledge into the parameters of pre-trained language models through direct parameter modifications. 
Existing approaches can generally be categorized into two paradigms: locate-then-edit methods, represented by ROME \citep{meng2022locating}, and hypernetwork-based methods, represented by MEND \citep{mitchell2021fast}, which learn parameter update directions through additional training. 
In this work, we primarily focus on the former approach.

Locate-then-edit methods first localize the transformer layers and hidden representations associated with target knowledge, typically through causal tracing \citep{meng2022locating,meng2022mass,fang2025alphaedit}, and then modify the corresponding parameters to steer the model toward the desired outputs.
Among these approaches, \citet{meng2022locating} introduce ROME, a single-layer editing method designed for individual factual associations.
Subsequently, \citet{meng2022mass} present MEMIT, which extends this framework to batch editing of multiple facts. 
Building upon MEMIT, \citet{fang2025alphaedit} propose AlphaEdit, which incorporates a null-space constraint to reduce interference from parameter updates and mitigate undesired impacts on the model's existing knowledge and capabilities.

Recent studies have further extended model editing to LLM personalization. \citet{huang2025towards} use clustering to augment each target preference with multiple semantically similar samples, in order to improve the editing accuracy of individual preferences. 

However, scaling model editing to a large number of personalized preferences remains a challenge. Prior studies have shown that applying model editing to a large number of samples may lead to forgetting and downstream performance degradation \citep{gupta2024model}.
This issue is particularly relevant in LLM personalization, where a user may have a large number of personalized preferences. 
Directly incorporating all user interaction data into model editing not only incurs substantial computational overhead, but may also worsen catastrophic forgetting and reduce the scalability of model editing.

Therefore, our work considers a user with multiple personalized preferences and uses hidden-representation clustering to select a compact set of representative preferences before editing. 
This strategy aims to reduce editing cost while maintaining personalization effectiveness and overall model performance. 
Although both approaches employ clustering, their purposes are different: \citet{huang2025towards} use clustering for preference augmentation, whereas we use it to select representative preferences under a limited editing budget.


\section{\textsc{PersonaEdit}}

\subsection{Editing Sample Selection Strategy}

Prior work has shown that the effects of model editing can transfer to related inputs, suggesting that editing effects are not limited to the edited instance
\citep{meng2022locating,meng2022mass,fang2025alphaedit}. 
In addition, existing studies have observed that editing facts with similar subjects may lead to interference effects, suggesting potential dependencies among editing outcomes across different inputs \citep{dong2025memit}.
Motivated by these findings, we hypothesize that similarity in hidden representations may be associated with partially shared internal computational pathways, which in turn may affect the locality of editing effects.

The overall procedure can be formulated as the following three stages.

\subsubsection{Extract Hidden Representations}
For each training sample, we feed the input prompt containing the subject $s_i$ into a baseline language model and extract the hidden state of the last token corresponding to the subject from an intermediate layer used in model editing methods, resulting in a hidden representation vector $\mathbf{h}_i \in \mathbb{R}^d$. The choice of this layer follows the design settings in prior model editing methods (e.g., AlphaEdit \citep{fang2025alphaedit}), ensuring that the extracted hidden representations are aligned with the model layer on which subsequent editing operations are performed.

\subsubsection{Clustering via K-Means}

To aggregate information and reduce potential overlap in computational pathways, we apply K-means clustering to the feature vectors $\{\mathbf{h}_i\}_{i=1}^N$, partitioning them into $K$ disjoint clusters by minimizing the intra-cluster variance:
\begin{equation}
    \arg\min_{\mathcal{C}} \sum_{k=1}^K \sum_{\mathbf{h}_i \in C_k} \|\mathbf{h}_i - \boldsymbol{\mu}_k\|^2
\end{equation}
where $\boldsymbol{\mu}_k$ denotes the centroid of cluster $C_k$.
Through this clustering process, input prompts with similar hidden representations are grouped into the same cluster, reflecting potentially shared local computational structures within the model.

\subsubsection{Proportional Stratified Sampling}

To maintain overall coverage of the representation space under a limited number of selected samples while reducing sampling bias, we perform proportional stratified sampling within each cluster.
Given the full training pool $\mathcal{D}$ containing $N = 300$ samples, we consider different settings for the number of selected samples $M$. The number of samples assigned to each cluster $C_k$ is defined as:
\begin{equation}
    n_k = \frac{|C_k|}{N} \cdot M
\end{equation}

In practice, we round $n_k$ to integers and further adjust the allocation to ensure that $\sum_k n_k = M$.
We then rank samples within each cluster based on their Euclidean distance to the centroid $\boldsymbol{\mu}_k$ and select the top-$n_k$ nearest samples, forming the final editing set $\mathcal{D}_{\text{edit}}$.

\section{Experiments}

In this section, we compare model editing with retrieval-based and prompt-based personalization methods to evaluate their relative effectiveness and potential complementarity in personalization tasks.

\subsection{Dataset Preprocessing and Subsampling}
\label{sec:sampling}

We conduct experiments using OpinionQA \citep{santurkar2023whose}, which contains longitudinal user-level question-answering data collected across 15 interaction waves, capturing individuals’ opinions on various social and political topics.
Due to its large scale, we select users who participate in all 15 interaction waves and further perform stratified sampling based on the performance of the original model. 
Specifically, we sample 10 users from each of the top-performing group (top 10), middle group (middle 10), and lowest-performing group (bottom 10), resulting in a total of 30 users and 31,017 interaction instances, to ensure diversity in user characteristics.

In the dataset construction, we consider 30 respondents whose question sets partially overlap. We therefore sample 300 questions from those commonly answered by all users to form the training set for subsequent clustering, while the remaining questions are used as the test set for personalized evaluation.
In the test set, most users have over 700 test samples. Overall, the final dataset consists of 9,000 training samples and 22,017 test samples.

After selecting the experimental subset, we use the Gemini 3 Flash model to identify the subject entity from each original OpinionQA question. 
We then construct structured subject–relation–object (s, r, o) triplets by treating the remaining question text as the relation and the respondent's original answer as the object, in order to support locate-then-edit based model editing methods \citep{meng2022locating}.
For example, the question “How much, if at all, do you worry about the following happening to you? Being the victim of a terrorist attack” with the response “Worry a little” is converted into (victim of a terrorist attack, How much, if at all, do you worry about the following happening to you? Being the, Worry a little). 
The subject is identified from the original question, while the remaining question text and the original response form the relation and object, respectively.

Additionally, to evaluate the generalization ability of model editing under semantic rephrasing, we also generate corresponding paraphrased queries for each question.
The prompts used for this transformation are provided in  Appendix~\ref{appendix:prompts}.

\subsection{Experimental Setup}
We compare the proposed method against several baselines, including Vanilla, Few-shot, Persona, All-info (Persona + Few-shot) and FERMI \citep{kim2025few}. 
The experiments are conducted using LLaMA 3.1 8B  \citep{meta2024llama3} under the k=18, training set size=200 setting.

In the few-shot setting, we retrieve the top-3 relevant samples for each task using BM25 and Contriever, and incorporate them into the prompt. 
In the persona setting, we use basic demographic information provided by OpinionQA \citep{santurkar2023whose} (e.g., gender, age, and income) as personalization conditions. 
The all-info setting combines both the top-3 retrieved samples and persona information in the prompt. 
FERMI adopts an iterative prompt optimization approach for inference \citep{kim2025few}.

\subsection{Main Personalization Results}

We first evaluate personalization performance using accuracy, where the final prediction is determined by matching the option in the model output. 
The experimental results are presented in Table~\ref{tab:opinionqa_respondent_analysis}. 
The results show that model editing achieves comparable or better performance than retrieval-based and prompt-based personalization methods. 
In particular, the combination of model editing and All-info achieves the best accuracy of 47.00\%, outperforming all baseline approaches. 
Respondent-level analysis further demonstrates consistent improvements across users, with gains observed for all 30 respondents under model editing.

\begin{table}[t]
\centering
\small
\begin{tabular}{lcccc}
\toprule
Method & Accuracy & Gain & \# Gain \\
\midrule
Vanilla                                 & 33.38\% & --      & -- \\
Few-shot$_{\text{BM25}}$                         & 43.00\% & 9.67\%  & 30 \\
Few-shot$_{\text{contriever}}$                      & 42.84\% & 9.49\%  & 30 \\
Persona                                 & 41.44\% & 8.05\%  & 30 \\
All info$_{\text{BM25}}$                           & 44.60\% & 11.28\% & 30 \\
All info$_{\text{contriever}}$                       & 44.51\% & 11.15\% & 30 \\
FERMI                                  & 34.52\% & 1.16\%  & 16 \\
\midrule
PersonaEdit                           & 44.67\% & 11.31\% & 30 \\
 + persona                 & 45.35\% & 12.01\% & 30 \\
 + Few-shot$_{\text{BM25}}$            & \underline{45.73\%} & \underline{12.39\%} & 30 \\
 + All info$_{\text{BM25}}$  & \textbf{47.00\%} & \textbf{13.65\%} & 30 \\
\bottomrule
\end{tabular}
\caption{Evaluation results on the OpinionQA dataset. 
Bold and underlined values indicate the best and second-best performance, respectively. 
``Gain'' denotes the average accuracy improvement per respondent compared to the Vanilla baseline, and ``\# Gain'' indicates the number of respondents with performance gains.}
\label{tab:opinionqa_respondent_analysis}
\end{table}

\paragraph{Discussion on FERMI Reproduction.}

As shown in Table~\ref{tab:opinionqa_respondent_analysis}, FERMI achieves lower performance than the other personalization baselines.
We believe this discrepancy may stem from several differences between our setup and the original FERMI framework.

First, the original FERMI study primarily evaluates its method using larger-scale models, such as ChatGPT and Llama 2 70B, whereas our experiments are conducted mainly on the smaller Llama 3.1 8B Instruct model. 
During implementation, we observe that smaller models are more likely to exhibit unstable output formatting behavior during iterative prompt optimization. 
In particular, the optimized prompts occasionally fail to effectively constrain the model to select from predefined options, causing the model to generate verbose free-form responses rather than directly outputting the corresponding option, which negatively affects the final answer matching accuracy.

\begin{table}[t]
\centering
\small
\setlength{\tabcolsep}{5pt}
\begin{tabular}{lcccc}
\toprule
Prompt & Vanilla & PersonaEdit & Gain & \# Gain \\
\midrule
w/o prompt & 33.38\% & 44.67\% & 11.31\% & 30 \\
FS$_{\text{(BM25)}}$  & 43.00\% & 45.35\% & 2.72\%           & 25 \\
Persona                     & 41.44\% & 45.73\% & 3.96\% & 25 \\
All info$_{\text{(BM25)}}$  & 44.60\% & 47.00\% & 2.51\%           & 25 \\
\bottomrule
\end{tabular}
\caption{Comparative analysis of Vanilla and PersonaEdit across different prompt augmentation strategies on the OpinionQA dataset. 
Bold and underlined values indicate the best and second-best performance within columns. 
FS denotes few-shot prompting.
``Gain'' denotes the average accuracy improvement per respondent brought by model editing, and ``\# Gain'' indicates the number of respondents who achieved gains.}
\label{tab:prompt_setting_comparison}
\end{table}

In addition, to ensure a consistent evaluation protocol across all personalization methods, our experimental setting requires every method to generate full answer content instead of producing a single option token (e.g., A/B/C/D). 
However, the original FERMI framework is primarily evaluated under a constrained option prediction setting. 
Under full-answer generation settings, FERMI's prompt optimization process appears to have greater difficulty maintaining output formatting, particularly when applied to smaller models.

A second possible reason stems from differences in the experimental data setting. 
The original FERMI study mainly focuses on small-scale personalization scenarios, where each user typically contains only a small number of training and testing samples (e.g., 20 training samples and 30 testing samples per user). 
In contrast, our experimental setting contains approximately 200 training samples and approximately 700 testing samples for each user. 
Since the data scale and personalization setting differ substantially from those used in the original FERMI study, the method design of FERMI may not directly generalize to larger-scale personalization scenarios.

Finally, because the official implementation of FERMI has not been released, our experiments are based on a reimplementation derived from the methodological descriptions provided in the original paper. 
As a result, implementation details may differ from those of the original system.

\subsection{Effect of Prompt Augmentation}

Beyond the comparison with existing personalization baselines, we further investigate whether model editing can be effectively combined with retrieval-based and prompt-based augmentation strategies.
As shown in Table~\ref{tab:prompt_setting_comparison}, the edited model consistently outperforms the non-edited model under all prompt settings, with statistically significant improvements. 
Respondent-level analysis reveals that 25 out of 30 respondents exhibit performance gains when using the edited model.

\paragraph{Comparison with Frontier Closed-Source Models.}
To further contextualize prompting-based personalization, we evaluate GPT-5.4 and GPT-5.4 mini under the same prompting settings. 
As shown in Table~\ref{tab:gpt_comparison}, under the strongest prompting setting (All-info$_{\text{(BM25)}}$), GPT-5.4 and GPT-5.4 mini achieve 51.51\% and 49.72\% accuracy, respectively, compared with 47.00\% for PersonaEdit.
While stronger closed-source models achieve higher accuracy, prompting-based personalization requires user information to be repeatedly provided during inference, or risks persona forgetting in multi-term interactions \citep{huang2025towards}. In contrast, model editing directly incorporates user-specific preferences into model parameters for open-weight models.

\begin{table}[t]
\centering
\small
\setlength{\tabcolsep}{5pt}
\begin{tabular}{lcccc}
\toprule
Prompt  & PersonaEdit & GPT-5.4 mini & GPT-5.4 \\
\midrule
w/o prompt  & 44.67\% & 43.89\% & 37.63\% \\
Few-shot$_{\text{(BM25)}}$  & 45.35\% & 47.64\% & 48.39\% \\
Persona  & 45.73\% & 47.01\% & 49.67\% \\
All info$_{\text{(BM25)}}$  & 47.00\% & 49.72\% & 51.51\% \\
\midrule
Total Cost & -- & \$9.63 & \$57.58 \\
\bottomrule
\end{tabular}
\caption{Comparison of personalization performance between PersonaEdit and closed-source GPT-5.4 mini and GPT-5.4 models.}
\label{tab:gpt_comparison}
\end{table}

\subsection{Ordinal Evaluation}

Additionally, since some questions in OpinionQA exhibit ordinal relationships among options, accuracy alone cannot capture the distance between predictions and ground-truth labels. 
Therefore, for these cases, we additionally adopt macro-averaged MAE (MMAE) \citep{baccianella2009evaluation} and the Closeness Evaluation Measure (CEM) \citep{amigo2020effectiveness} as evaluation metrics.

Macro-averaged MAE measures the average distance between predicted and ground-truth labels. 
CEM is an ordinal classification evaluation metric proposed by \citet{amigo2020effectiveness}, designed to account for the informational closeness between ordinal classes.

The results are shown in Table ~\ref{tab:opinionqa_ordinal_analysis}. 
In terms of accuracy, the model editing approach significantly outperforms the other methods. For the MMAE metric, editing-based approaches also achieve significantly better performance than the baselines. 
In terms of CEM, Model Editing with Few-shot$_{\text{(BM25)}}$ significantly outperforms the other approaches as well. 
Detailed statistical significance test results for this section are provided in  Appendix~\ref{appendix:significance_test}.

\begin{table}[t]
\centering
\small
\begin{tabular}{lccc}
\toprule
Method & Accuracy & MMAE ($\downarrow$) & CEM ($\uparrow$) \\
\midrule
Vanilla                                   & 35.48\% & 1.0728 & 0.5356 \\
Few-shot$_{\text{(BM25)}}$                & 41.47\% & 1.0842 & 0.5370 \\
Few-shot$_{\text{(Contriever)}}$          & 40.33\% & 1.0802 & 0.5381 \\
Persona                                   & 38.52\% & 1.0952 & 0.5377 \\
All info$_{\text{(BM25)}}$                & 42.50\% & 1.1067 & 0.5350 \\
All info$_{\text{(Contriever)}}$          & 40.97\% & 1.0986 & 0.5366 \\
FERMI                                     & 34.56\% & 1.1237 & 0.5232 \\
\midrule
PersonaEdit                             & 43.00\% & \textbf{1.0669} & \underline{0.5382} \\
+ Persona                   & 42.84\% & 1.0744          & 0.5379 \\
+ Few-shot$_{\text{(BM25)}}$ & \underline{43.76\%} & 1.0825          & \textbf{0.5385} \\
+ All info$_{\text{(BM25)}}$  & \textbf{44.95\%} & \underline{1.0733} & 0.5371 \\
\bottomrule
\end{tabular}
\caption{Evaluation results on the OpinionQA dataset focusing on questions with ordinal options. Bold and underlined values indicate the best and second-best performance within each column, respectively. $\uparrow$ indicates higher values are better, while $\downarrow$ indicates lower values are better.}
\label{tab:opinionqa_ordinal_analysis}
\end{table}

\section{Discussion}

We conduct a series of experiments and analyses to evaluate the effectiveness of the proposed hidden representation clustering-based sample selection method for personalized model editing tasks.

\begin{itemize}[nosep]
    \item \textbf{RQ1:} How effective is model editing for personalization across different model architectures and editing methods?
    \item \textbf{RQ2:} How do different sample selection strategies affect personalization performance in model editing?
    \item \textbf{RQ3:} How do the number of clusters and the sample selection size affect personalization performance?
    \item \textbf{RQ4:} What factors determine the effectiveness of model editing for personalization?
\end{itemize}

\subsection{Model Editing for Personalization (RQ1)}
This section evaluates the performance of LLaMA 3.1 8B Instruct and GPT-J-6B \citep{gpt-j} under three model editing methods (ROME, MEMIT, and AlphaEdit) on the personalized dataset constructed in the previous sections.

Following prior model editing studies \citep{fang2025alphaedit}, we adopt four evaluation criteria: efficacy, paraphrase, fluency, and locality. 
In the locality evaluation, we further use tasks from the General Language Understanding Evaluation (GLUE) benchmark \citep{wang2018glue}, including SST-2 \citep{socher2013recursive}, MRPC \citep{dolan2005automatically}, RTE \citep{bentivogli2009fifth}, and CoLA \citep{warstadt2019neural}, as well as natural language inference (NLI) tasks \citep{williams2018broad}. 
In addition, we include MMLU \citep{hendrycks2020measuring} as a supplementary benchmark to assess the model’s reasoning ability.

As shown in Table~\ref{tab:editing_methods}, across both LLaMA 3.1 8B Instruct and GPT-J-6B, MEMIT and AlphaEdit consistently outperform ROME on most metrics, including efficacy, paraphrase, and fluency.
A further comparison between MEMIT and AlphaEdit reveals that the two methods achieve comparable performance, while exhibiting different advantages across model architectures. 
On LLaMA 3.1 8B Instruct, AlphaEdit significantly outperforms MEMIT in paraphrase robustness (paired Wilcoxon test, p < 0.001). 
On GPT-J-6B, AlphaEdit shows significant improvements over MEMIT in multiple metrics, including efficacy, fluency, and locality (paired Wilcoxon test, p < 0.005).

\begin{table}[t]
\centering
\small
\setlength{\tabcolsep}{5pt}
\begin{tabular}{llcccc}
\toprule
Model & Method & Eff. & Para. & Flu. & Loc. \\
\hline
\multirow{4}{*}{LLaMA}
& Vanilla   & 39.65 & 41.02 & 641.15 & 62.85 \\
& ROME      & 39.28 & 40.83 & \textbf{634.62} & 62.51 \\
& MEMIT     & \textbf{89.67} & \underline{64.68} & \underline{631.45} & \textbf{62.95} \\
& AlphaEdit & \underline{89.47} & \textbf{66.33} & 630.93 & \underline{62.88} \\
\hline
\multirow{4}{*}{GPT-J}
& Vanilla   & 27.18 & 23.92 & 621.41 & 45.45 \\
& ROME      & 28.32 & 23.68 & \textbf{618.69} & \textbf{45.53} \\
& MEMIT     & \underline{95.75} & \underline{47.30} & 609.33 & 43.57 \\
& AlphaEdit & \textbf{97.35} & \textbf{47.83} & \underline{610.48} & \underline{44.03} \\
\bottomrule
\end{tabular}
\caption{Comparison of model editing methods across different base models.
Locality is reported as the average performance across GLUE and MMLU benchmarks.
Eff., Para., Flu., and Loc. denote efficacy, paraphrase, fluency, and locality, respectively.
Bold and underlined values indicate the best and second-best performance within each model group, respectively (excluding the Vanilla baseline for Fluency ranking).}
\label{tab:editing_methods}
\end{table}

\begin{table*}[t]
\centering
\small
\begin{tabular}{lccccccc}
\toprule
Evaluation Set & Mean $\Delta$ Acc. & Std & Min $\Delta$ & Max $\Delta$ & \# Improved & \# Degraded & \# Unchanged \\
\hline
Edit set              & +49.58 & 10.09 & +30.83 & +61.67 & 10 & 0 & 0 \\
Same-topic hold-out   & +1.60  & 11.07 & -11.76 & +22.58 & 3  & 4 & 3 \\
Other-topic hold-out  & +3.54  & 2.85  & -1.32  & +7.79  & 9  & 1 & 0 \\
\bottomrule
\end{tabular}
\caption{Editing influence across different hold-out settings under topic-based sample selection. Mean, standard deviation, minimum, and maximum $\Delta$ Acc. are computed across 10 respondents. \# Improved, \# Degraded, and \# Unchanged indicate the number of respondents (N=10).}
\label{tab:topic_influence}
\end{table*}

\begin{figure*}[t]
\centering 
\includegraphics[width=\textwidth]{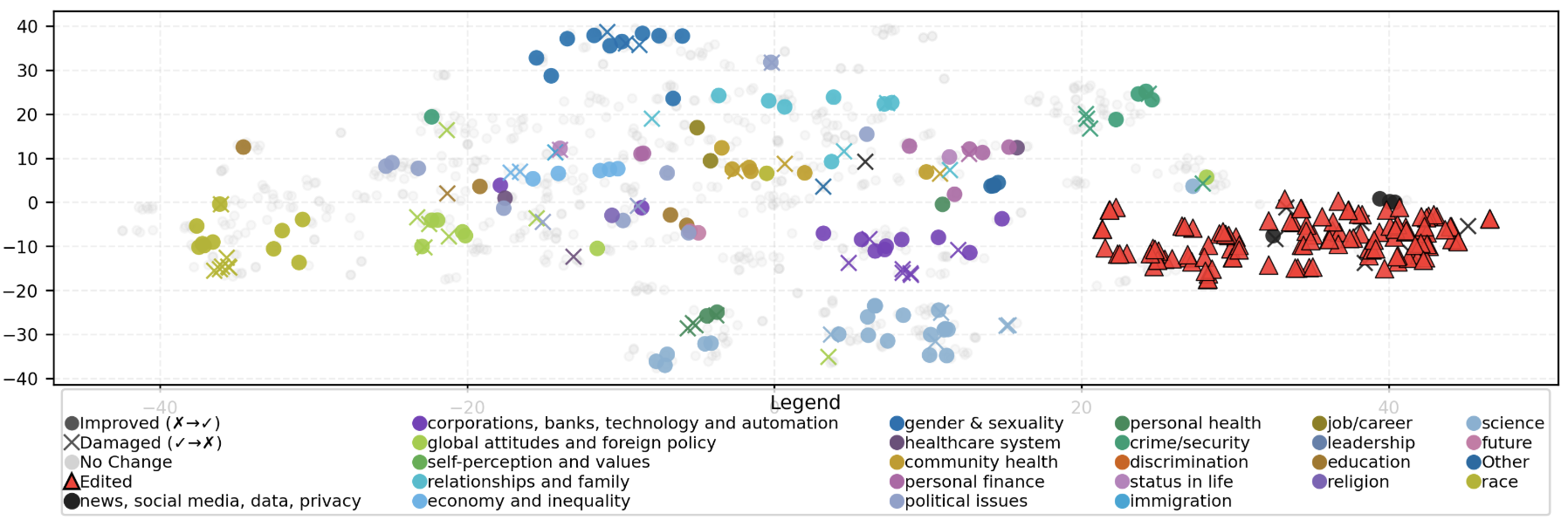} 
\caption{Semantic similarity visualization illustrating the transfer effects of model editing.
Red triangles denote the edited samples used for model editing, while black markers represent other samples from the same semantic topic category.
Circles indicate samples whose predictions improved after editing, whereas crosses denote samples whose predictions became incorrect after editing.}
\label{fig:semantic_similarity}  
\end{figure*}

Furthermore, in the locality evaluation, the differences among methods are relatively small, indicating that model editing approaches can improve the accuracy of target knowledge while having limited impact on the model’s original language and reasoning capabilities, without degrading the model’s inherent performance.
Detailed locality evaluation results are provided in Appendix~\ref{appendix:locality}.

\begin{figure*}[t]
\centering 
\includegraphics[width=\textwidth]{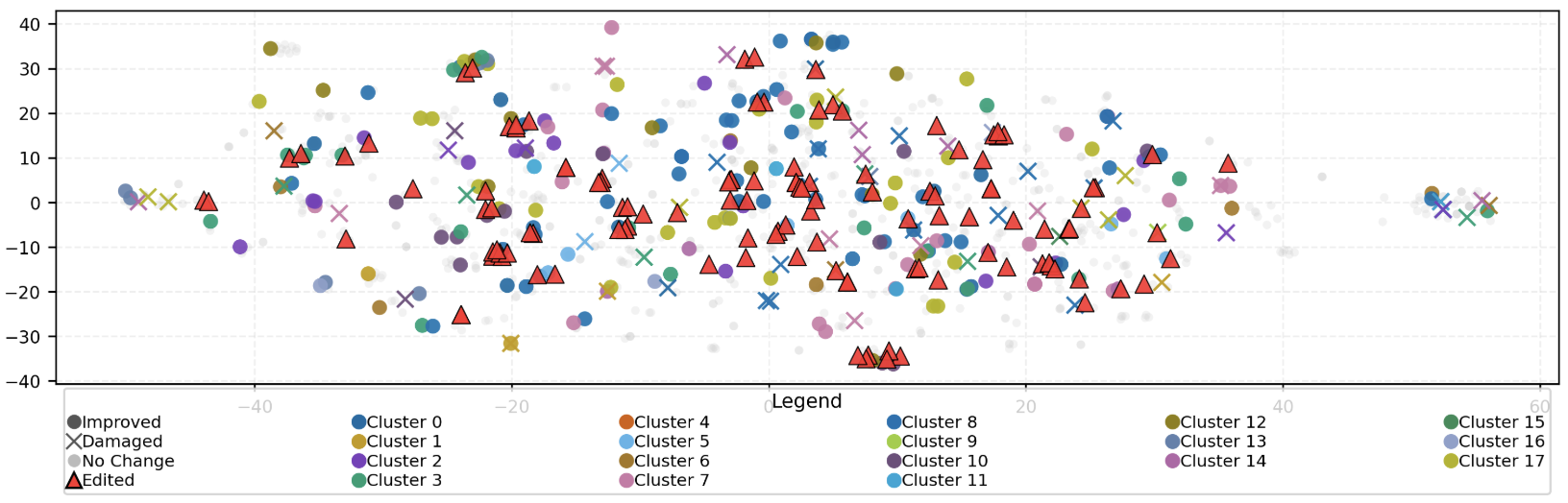} 
\caption{Hidden representation visualization illustrating the transfer effects of model editing.
Red triangles denote the edited samples used for model editing.
Circles indicate samples whose predictions improved after editing, whereas crosses denote samples whose predictions became incorrect after editing.
Different colors represent clusters constructed from hidden representation similarity.}
\label{fig:hidden_cluster}  
\end{figure*}

\subsection{Effect of Sample Selection Strategies (RQ2)}
In this section, we investigate how different sample selection strategies affect personalization performance in model editing, with a particular focus on human-annotated topic-based selection.

We use the topic annotations provided in the original OpinionQA dataset, which categorizes questions into 23 social and political topics with varying numbers of samples per topic, and randomly select 10 respondents who participated in all 15 survey waves.
For each setting, we perform model editing using 120 samples from the topic containing the largest number of questions and evaluate the edited models on both same-topic and cross-topic hold-out samples.

As shown in Table~\ref{tab:topic_influence}, editing effects vary substantially across respondents even when both editing and evaluation samples are drawn from the same topic, with both performance improvements and degradations observed. 
In contrast, the cross-topic setting exhibits relatively more consistent behavior. 
These findings suggest that human-annotated topics may not reliably capture the generalization behavior of model editing.
More detailed per-respondent results are provided in  Appendix~\ref{appendix:topicresults}.

To further analyze this phenomenon, we examine the relationship between editing influence and the cosine similarity between edited and hold-out samples using embeddings generated by all-MiniLM-L6-v2. We further visualize the embedding distribution using t-SNE. 
As shown in Figure~\ref{fig:semantic_similarity}, there is no consistent monotonic relationship between semantic similarity and editing influence.
Although some affected samples exhibit higher semantic similarity to the edited samples, samples with low similarity can also undergo substantial changes. 
For example, several samples from semantically distant categories, such as \textit{race}, are also affected after editing, suggesting that editing transferability may not align well with human-annotated semantic topics.
Compared with the hidden representation clustering visualization in Figure~\ref{fig:hidden_cluster}, these observations suggest that editing influence may be better characterized by hidden representation similarity than by human-annotated semantic similarity.

Motivated by this observation, we adopt hidden representation clustering as a sample selection strategy, grouping samples according to similarity in the model’s internal representations to better capture editing transfer behavior.

To further assess the effectiveness of hidden representation clustering, we compare it with random sample selection using the same preference pool as in our main experiments (N=300) and an editing size of M=20. 
For random selection, we use two different random seeds and report the average performance. 
Our clustering-based selection achieves higher test accuracy than random sampling (0.3823 vs. 0.3756), and a paired significance test across respondents indicates that the improvement is statistically significant (p=0.0451).

\subsection{Clustering and Sample Size Effects (RQ3)}

\begin{figure}[t]
\centering
\includegraphics[width=7cm]{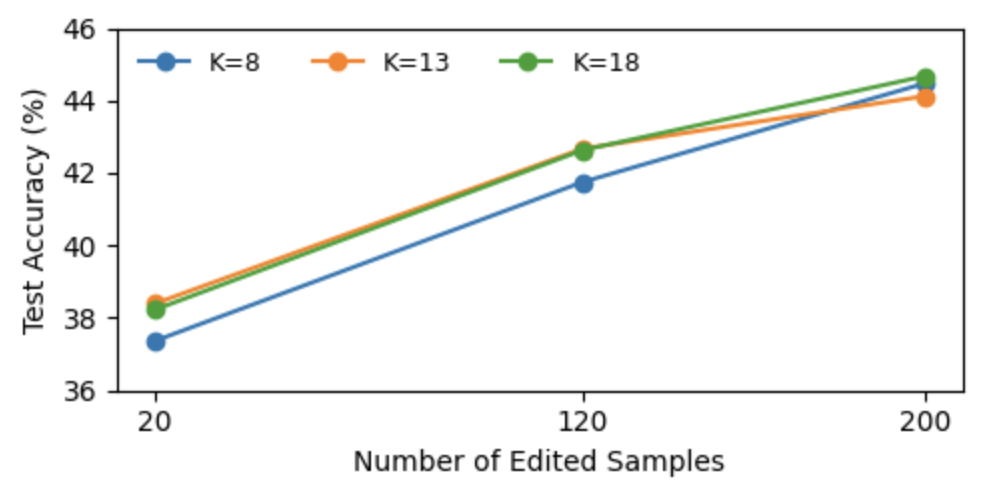}
\caption{
Test accuracy improvement under different clustering hyperparameters $K$ and editing data sizes.
}
\label{fig:k_sample_accuracy}
\end{figure}

In this section, we analyze the effects of different K values (8, 13, and 18) and editing data sizes (20, 120, and 200 samples).
Figure~\ref{fig:k_sample_accuracy} illustrates the test accuracy under different settings.

According to the results, the most influential factor affecting model editing performance is the amount of editing data. 
Regardless of the choice of K, performance on the test set consistently improves as the number of editing samples increases, yielding approximately 5\%--7\% gains overall.

However, the performance improvement begins to plateau as the editing data size increases. 
In particular, increasing the number of editing samples from 120 to 200 yields only marginal gains (approximately 1\%--2\%), suggesting that a moderate number of representative samples is sufficient to preserve most of the personalization performance.

In contrast, the choice of K does not substantially affect editing performance.
Under the same editing data size, changing K from 8 to 18 results in only minor fluctuations in accuracy (mostly within 1\%--2\%), without showing a consistent positive or negative trend.
Detailed editing-related evaluation metrics are provided in Appendix~\ref{appendix:editing_metrics}.

Finally, we conduct an additional analysis to examine whether the effectiveness of sample selection depends on the size of the preference pool. 
We evaluate our method with a smaller pool of N=100, while keeping the editing size fixed at M=20. 
Our method achieves comparable test accuracy under the smaller pool setting (0.3770 for N=100 vs. 0.3823 for N=300), suggesting that the selection strategy remains effective when fewer user preferences are available.

\subsection{Factors of Editing Effectiveness (RQ4)}

In this section, we analyze the effects of model editing from two perspectives: (1) how the alignment between users and the original model influences editing performance, and (2) how the consistency of user responses within the same cluster affects model accuracy.

\subsubsection{Effect of User-Model Alignment}

First, we analyze how user-model alignment influences the effectiveness of model editing. As described in Section~\ref{sec:sampling}, we conduct editing experiments on the top 10, middle 10, and bottom 10 users ranked by original model accuracy.
Table~\ref{tab:accuracy_groups} presents the average accuracy before and after editing for each group.

The results show that all three groups achieve noticeable improvements after model editing. 
Among them, the Low Accuracy group exhibits the largest improvement, while the High Accuracy group shows relatively smaller gains. 
This suggests that model editing tends to provide greater benefits when the alignment between user preferences and the original model is relatively low.

\begin{table}[t]
\centering
\small
\begin{tabular}{lccc}
\toprule
Group & Before & After & Improvement \\
\midrule
Low Accuracy  & 0.280 & 0.401 & +0.121 \\
Mid Accuracy  & 0.335 & 0.447 & +0.113 \\
High Accuracy & 0.387 & 0.492 & +0.105 \\
\bottomrule
\end{tabular}
\caption{Average editing performance before and after editing across different accuracy groups.}
\label{tab:accuracy_groups}
\end{table}

\subsubsection{Effect of Intra-Cluster Consistency}

In this section, we investigate whether user response consistency within hidden representation clusters influences personalization performance after model editing.
For each user response, we extract token embeddings from the language model output embedding space (i.e., \texttt{lm\_head}), average them to obtain the answer representation, and apply $L_2$ normalization.
We then compute the average pairwise cosine similarity among all answer representations within each cluster as the intra-cluster answer consistency score, and compute a weighted average based on the number of questions in each cluster to obtain a respondent-level consistency score.

We further examine the relationship between this consistency score and the accuracy improvement achieved through model editing using Spearman rank correlation analysis. 
The results reveal a positive correlation between response consistency and personalization improvement, with statistically significant correlation on the training set ($\rho = 0.4218, p = 0.0202$) and a moderate positive trend on the test set ($\rho = 0.2984, p = 0.1090$). 
These results suggest that users with more consistent response patterns may benefit more from model editing personalization.
Conversely, users with more heterogeneous or conflicting preferences may be more challenging to personalize, as the selected representative samples may not adequately capture the diversity of their preferences.


\section{Conclusion}

Recent studies have begun to explore model editing for personalization.
In this paper, we investigate how sample selection strategies influence editing generalization and personalization performance.
Experiments on OpinionQA show that hidden representation similarity provides a more effective basis for editing sample selection than human-annotated semantic topics, while the proposed clustering strategy reduces required editing samples while preserving most personalization performance.
We further observe that retrieval-based prompt augmentation and model editing complement each other in personalization tasks, and that user-level characteristics influence editing effectiveness.

Despite these promising results, the current study is limited to a single benchmark and restricted editing settings. Future work may further investigate editing transferability across broader personalization scenarios.

\section*{Limitations}

Although the experimental results demonstrate the effectiveness of the proposed sample selection strategy for personalization-oriented model editing, several limitations remain. 
First, our evaluation is primarily based on OpinionQA and is conducted on a subset of 30 users, with a focus on social and political topics. 
Moreover, personalization is formulated as a discrete option prediction task.
As a result, the current evaluation setting does not fully capture more open-ended personalization scenarios, such as long-form dialogue generation, continual interaction, or dynamic shifts in user preferences over time.

Second, although hidden representation-based clustering empirically improves editing sample selection, we do not extensively compare it with other sample selection strategies. 
In particular, methods that cluster samples using external text embeddings could provide alternative selection signals. 
Comparing against a broader range of selection strategies would help better assess the effectiveness and advantages of model-internal representations for representative sample selection.

Finally, our experiments are conducted on a limited set of open-weight language models. 
Since editing behavior may vary across model architectures and training settings, the generalizability of the current findings remains limited.

In addition, our study evaluates personalized model editing independently for each user under a controlled experimental setting and does not specify a concrete deployment architecture. 
In practical deployment, maintaining a separately edited model or model delta for each user may introduce additional challenges, including update frequency, edit rollback, privacy protection, and cumulative interference from repeated edits.  
Personalization through model editing may also introduce potential risks related to privacy leakage, behavioral manipulation, or unintended amplification of user biases if deployed without appropriate safeguards. 
Addressing these issues is beyond the scope of the current study and requires further investigation.
Future work should further investigate editing generalization in broader personalization settings, practical deployment strategies, and safer and more controllable personalization mechanisms for real-world use.

\section*{Acknowledgments}

This research was partially supported by National Science and Technology Council, Taiwan, under grants NSTC 114-2221-E-A49-057-MY3, NSTC 114-2221-E-002-070-MY3 and NSTC 115-2634-F-002-012-, and Ministry of Education (MOE) in Taiwan, under grants 115L900901.


\bibliography{custom}

\appendix
\section{Implementation Details and Dataset Information}

Our implementation is primarily based on the publicly available AlphaEdit framework.  
Experiments are conducted on NVIDIA A6000 and H100 GPUs using LLaMA 3.1 8B Instruct and GPT-J-6B under multiple model editing settings.
The runtime of the editing pipeline on an H100 GPU is summarized in Table~\ref{tab:cost}.

\begin{table}[t]
\centering
\caption{Runtime of the model editing pipeline on an NVIDIA H100 GPU. 
The AlphaEdit projection computation is performed once and reused across editing experiments.}
\label{tab:cost}
\begin{tabular}{lc}
\toprule
Component & Runtime \\
\midrule
AlphaEdit projection (one-time) & $\sim$25 hours \\
Editing 200 samples & $\sim$2 hours \\
\bottomrule
\end{tabular}
\end{table}

Unless otherwise specified, the editing hyperparameter settings follow those used in prior model editing studies.

We use the publicly released OpinionQA dataset, which is constructed from the American Trends Panel (ATP) surveys conducted by Pew Research. 
The dataset consists of 15 survey waves covering a wide range of social and political topics, including guns, race, and other public opinion topics.
Each respondent is associated with persona-related attributes such as sex, age, and education, together with coarse-grained and fine-grained topic annotations for each question.

An example question from OpinionQA is:
\textit{"How safe, if at all, would you say your local community is from crime? Would you say it is (Options: Very safe, Somewhat safe, Not too safe, Not at all safe, Refused)"}

All datasets, models, and model editing frameworks used in this work are publicly available and used under their respective licenses and intended research usage conditions. 
Details regarding data collection procedures and participant consent are described in the original OpinionQA dataset paper.
The dataset does not include personally identifying information, and we do not collect or process any additional personal information.

\section{Prompt Details}
\label{appendix:prompts}

Previous studies on model editing commonly represent knowledge in the form of $(s, r, o)$ triplets, where $s$, $r$, and $o$ denote the subject, relation, and object, respectively.
This representation is adopted because prior locate-then-edit studies have shown that editing based on the hidden representation of the last token of the subject generally achieves superior performance \citep{meng2022locating}.

Therefore, in this study, we first transform each question into an (s,r,o)-style representation and identify the corresponding subject entity in each question. 
Specifically, we use Gemini 3 flash to automatically extract the subject for each question.

The prompt used for subject extraction is provided below.

\begin{tcolorbox}[breakable,
 colback=gray!5,colframe=black,boxrule=0.5pt,
 arc=2pt,left=4pt,right=4pt,top=4pt,bottom=4pt]
\footnotesize

\textbf{Task:} Extract a ``Continuous Discriminative Subject'' for Model Editing.\\

\textbf{Question:} ``\{question\}''\\

\textbf{Rules:}
\begin{itemize}
    \item No pronouns such as ``I'', ``you'', or ``my''.
    \item Include semantic context (e.g., gender or race) when relevant.
    \item The subject must be a continuous span appearing verbatim in the question.
    \item Avoid leading or trailing filler phrases.
    \item Ensure the subject is uniquely identifiable.
\end{itemize}

\textbf{Examples:}
\begin{itemize}
    \item ``How safe ... is your local community from crime?'' $\rightarrow$ local community
    \item ``... treated unfairly because of your gender ...'' $\rightarrow$ treated unfairly because of your gender
    \item ``Do you now smoke cigarettes?'' $\rightarrow$ now smoke cigarettes
\end{itemize}

\textbf{Output:} \texttt{\{"subject": "..."\}}

\end{tcolorbox}

\begin{table*}[t]
\centering
\small
\begin{tabular}{llcccccc}
\toprule
Model & Method & SST & MRPC & CoLA & RTE & NLI & MMLU \\
\midrule
\multirow{4}{*}{LLaMA 3.1 8B}
& Vanilla   & \underline{95.99}$_{{\pm}0.00}$ & \textbf{69.81}$_{{\pm}0.00}$ & 69.38$_{{\pm}0.00}$ & 22.77$_{{\pm}0.00}$ & \underline{61.62}$_{{\pm}0.00}$ & \textbf{57.52}$_{{\pm}0.00}$ \\
& ROME      & \textbf{96.13}$_{{\pm}1.41}$ & \underline{69.52}$_{{\pm}1.52}$ & 68.75$_{{\pm}0.62}$ & \textbf{22.82}$_{{\pm}1.61}$ & \textbf{62.44}$_{{\pm}1.49}$ & 55.42$_{{\pm}1.58}$ \\
& MEMIT     & 95.86$_{{\pm}0.34}$ & 66.10$_{{\pm}2.56}$ & \underline{76.37}$_{{\pm}2.80}$ & 21.18$_{{\pm}0.92}$ & 61.16$_{{\pm}0.92}$ & \underline{57.04}$_{{\pm}1.99}$ \\
& AlphaEdit & 95.79$_{{\pm}0.84}$ & 64.42$_{{\pm}1.30}$ & \textbf{78.29}$_{{\pm}2.09}$ & \underline{21.77}$_{{\pm}0.97}$ & 61.21$_{{\pm}1.66}$ & 55.78$_{{\pm}1.78}$ \\
\midrule
\multirow{4}{*}{GPT-J-6B}
& Vanilla   & 75.00$_{{\pm}0.00}$ & \underline{42.07}$_{{\pm}0.00}$ & \textbf{38.56}$_{{\pm}0.00}$ & \textbf{48.33}$_{{\pm}0.00}$ & 50.00$_{{\pm}0.00}$ & 18.75$_{{\pm}0.00}$ \\
& ROME      & \underline{75.08}$_{{\pm}0.28}$ & \textbf{43.04}$_{{\pm}0.86}$ & \textbf{38.56}$_{{\pm}0.00}$ & \underline{47.48}$_{{\pm}0.00}$ & 49.30$_{{\pm}0.46}$ & \textbf{19.74}$_{{\pm}0.55}$ \\
& MEMIT     & \textbf{75.49}$_{{\pm}0.64}$ & 35.84$_{{\pm}3.74}$ & 36.26$_{{\pm}1.18}$ & 43.97$_{{\pm}2.71}$ & \textbf{51.37}$_{{\pm}1.86}$ & 18.51$_{{\pm}1.33}$ \\
& AlphaEdit & 75.03$_{{\pm}0.96}$ & 37.32$_{{\pm}4.36}$ & 35.67$_{{\pm}1.18}$ & 46.37$_{{\pm}2.98}$ & \underline{50.40}$_{{\pm}1.84}$ & \underline{19.38}$_{{\pm}2.06}$ \\
\bottomrule
\end{tabular}
\caption{Detailed evaluation of locality performance across individual text classification and language understanding benchmarks. 
Bold and underlined values indicate the best and second-best performance within each model group, respectively.}

\label{tab:detailed_locality}
\end{table*}

\begin{table}[t]
\centering
\small
\begin{tabular}{lccc}
\toprule
User & Edit $\Delta$ & Same-Topic $\Delta$ & Cross-Topic $\Delta$ \\
\midrule
User 1  & +0.617 & +0.114 & +0.042 \\
User 2  & +0.375 & -0.125 & +0.062 \\
User 3  & +0.567 & +0.226 & +0.020 \\
User 4  & +0.458 & -0.067 & +0.009 \\
User 5  & +0.417 & -0.095 & -0.016 \\
User 6  & +0.508 & -0.040 & +0.013 \\
User 7  & +0.625 & +0.000 & +0.029 \\
User 8  & +0.608 & +0.111 & +0.034 \\
User 9  & +0.300 & +0.000 & -0.021 \\
User 10 & +0.450 & -0.176 & +0.061 \\
\bottomrule
\end{tabular}
\caption{Per-user editing performance changes on edited, same-topic, and cross-topic samples before and after editing.}
\label{tab:topics_results}
\end{table}

\begin{table*}[t]
\centering
\small
\begin{tabular}{lcccc}
\hline
\textbf{Compared Method} & \textbf{Baseline Mean} & \textbf{Method Mean} & \textbf{Improvement} & \textbf{Wilcoxon p-value} \\
\hline
PersonaEdit & 0.4464 & 0.4467 & +0.0003 & 0.940 \\
PersonaEdit + persona & 0.4464 & 0.4537 & +0.0073 & 0.280 \\
PersonaEdit + Few-shot$_{\text{BM25}}$ & 0.4464 & 0.4575 & +0.0111 & 0.058 \\
PersonaEdit + All info$_{\text{BM25}}$  & 0.4464 & 0.4702 & +0.0238 & 0.00013 \\
\hline
\end{tabular}
\caption{
User-level personalization accuracy across methods using BM25 retrieval.
Baseline Mean and Method Mean represent the mean user-level accuracy averaged across respondents.
All p-values are computed against the All info$_{\text{BM25}}$ baseline using paired Wilcoxon signed-rank tests.
}
\label{tab:bm25_user_level_accuracy}
\end{table*}

\begin{table*}[t]
\centering
\small
\begin{tabular}{lcccc}
\hline
\textbf{Compared Method} & \textbf{Baseline Mean} & \textbf{Method Mean} & \textbf{Improvement} & \textbf{Wilcoxon p-value} \\
\hline

PersonaEdit & 0.4451 & 0.4467 & +0.0016 & 0.657566 \\
PersonaEdit + persona& 0.4451 & 0.4537 & +0.0086 & 0.036921 \\
PersonaEdit + Few-shot$_{\text{BM25}}$& 0.4451 & 0.4575 & +0.0124 & 0.017897 \\
PersonaEdit + All info$_{\text{BM25}}$& 0.4451 & 0.4702 & +0.0251 & 0.00000514 \\
\hline
\end{tabular}
\caption{
User-level personalization accuracy across methods using Contriever retrieval.
Baseline Mean and Method Mean represent the mean user-level accuracy averaged across respondents.
All p-values are computed against the All info$_{\text{contriever}}$ baseline using paired Wilcoxon signed-rank tests.
}
\label{tab:contriever_user_level_accuracy}
\end{table*}

\begin{table*}[t]
\centering
\small
\begin{tabular}{lcccc}
\hline
\textbf{Compared Method} &
\textbf{Baseline Mean} &
\textbf{Method Mean} &
\textbf{Improvement} &
\textbf{Wilcoxon p-value} \\
\hline

All info$_{\text{BM25}}$ vs PersonaEdit
& 0.4254 & 0.4288 & +0.0034 & 0.626346 \\
All info$_{\text{BM25}}$ vs PersonaEdit + persona
& 0.4254 & 0.4271 & +0.0017 & 0.597808 \\
All info$_{\text{BM25}}$ vs PersonaEdit + Few-shot$_{\text{BM25}}$
& 0.4254 & 0.4368 & +0.0114 & 0.145999 \\
All info$_{\text{BM25}}$ vs PersonaEdit + All info$_{\text{BM25}}$
& 0.4254 & 0.4485 & +0.0231 & 0.000555 \\

\hline
\end{tabular}

\caption{
User-level ordinal accuracy comparison across methods.
Baseline Mean and Method Mean represent the mean user-level accuracy averaged across respondents.
All p-values are computed using paired Wilcoxon signed-rank tests against the corresponding retrieval baseline.
}
\label{tab:ordinal_accuracy_significance}

\end{table*}

\begin{table*}[t]
\centering
\small

\begin{tabular}{lcccc}
\hline
\textbf{Compared Method} &
\textbf{Baseline Mean} &
\textbf{Method Mean} &
\textbf{Improvement} &
\textbf{Wilcoxon p-value} \\
\hline

All info$_{\text{BM25}}$ vs PersonaEdit & 0.2936 & 0.2831 & +0.0105 & 0.073244 \\
All info$_{\text{BM25}}$ vs PersonaEdit + persona & 0.2936 & 0.2802 & +0.0133 & 0.004338 \\
All info$_{\text{BM25}}$ vs PersonaEdit + Few-shot$_{\text{BM25}}$ & 0.2936 & 0.2834 & +0.0102 & 0.007612 \\
All info$_{\text{BM25}}$ vs PersonaEdit + All info$_{\text{BM25}}$ & 0.2936 & 0.2764 & +0.0172 & $4.97 \times 10^{-5}$ \\
\hline
\end{tabular}
\caption{
User-level macro-averaged MAE comparison across methods.
Baseline Mean and Method Mean represent the mean user-level accuracy averaged across respondents.
All p-values are computed using paired Wilcoxon signed-rank tests against the corresponding retrieval baseline.
}
\label{tab:mae_significance}

\end{table*}

\begin{table*}[t]
\centering
\small

\begin{tabular}{lcccc}
\hline
\textbf{Compared Method} &
\textbf{Baseline Mean} &
\textbf{Method Mean} &
\textbf{Improvement} &
\textbf{Wilcoxon p-value} \\
\hline

All info$_{\text{BM25}}$ vs PersonaEdit & 0.5350 & 0.5383 & +0.0033 & 0.013663 \\
All info$_{\text{BM25}}$ vs PersonaEdit + persona & 0.5350 & 0.5380 & +0.0030 & 0.027741 \\
All info$_{\text{BM25}}$ vs PersonaEdit + Few-shot$_{\text{BM25}}$ & 0.5350 & 0.5385 & +0.0035 & 0.000111 \\
All info$_{\text{BM25}}$ vs PersonaEdit + All info$_{\text{BM25}}$ & 0.5350 & 0.5371 & +0.0021 & 0.004665 \\

\hline
\end{tabular}

\caption{
User-level CEM comparison across methods.
Baseline Mean and Method Mean represent the mean user-level accuracy averaged across respondents.
All p-values are computed using paired Wilcoxon signed-rank tests against the corresponding retrieval baseline.
}
\label{tab:cem_significance}

\end{table*}

Furthermore, paraphrase questions are commonly used in model editing evaluations to verify whether the model has genuinely retained the edited knowledge rather than merely memorizing a specific input formulation \citep{meng2022locating}. 
Therefore, we utilize Gemini 3 flash to rewrite the original questions and generate paraphrase questions, which are subsequently used as evaluation data for model editing.

The prompt used for paraphrase generation is provided below.

\begin{tcolorbox}[
    breakable,
    colback=gray!5,
    colframe=black,
    boxrule=0.5pt,
    arc=2pt,
    left=4pt,
    right=4pt,
    top=4pt,
    bottom=4pt
]
\footnotesize

\textbf{Task:} For the provided survey question (Q) and its options (Options), generate a structured JSON object for each option.\\

\textbf{Core Rules:}
\begin{itemize}
    \item \textbf{Variable Fields:} Only \texttt{target} and \texttt{input\_attribute} should change according to the option content.
    \begin{itemize}
        \item \texttt{target}: Must exactly match the text in the Options list.
        \item \texttt{input\_attribute}: A first-person self-description simulating the user. This must be written based on the target option.
    \end{itemize}

    \item \textbf{Fixed Fields:} For the same question, \texttt{question}, \texttt{question\_paraphrased}, and \texttt{implicit\_question} must remain identical across all generated objects.

    \item \textbf{Output Format:} Output only a pure JSON list \texttt{[...]}.
\end{itemize}

\textbf{JSON Field Definitions:}
\begin{itemize}
    \item \texttt{input\_attribute}: A first-person user statement simulated based on the target option.
    \item \texttt{question}: The original survey question.
    \item \texttt{question\_paraphrased}: A natural, conversational rephrasing of the question.
    \item \texttt{implicit\_question}: Explore the underlying cues or concrete lived experiences that lead the respondent to arrive at this conclusion. Do not directly ask about the respondent's choice. Instead, ask about a specific situation or feeling that would allow the choice to be inferred indirectly.
    \item \texttt{target}: The exact option text from the list.
\end{itemize}

\textbf{Question:} \texttt{\{question\}}\\
\textbf{Options:} \texttt{\{options\}}

\end{tcolorbox}

\section{Locality Results}
\label{appendix:locality}

This section provides detailed locality evaluation results across six benchmarks: SST, MRPC, CoLA, RTE, NLI, and MMLU.
Table~\ref{tab:detailed_locality} presents the per-task performance for both LLaMA 3.1 8B Instruct and GPT-J-6B models, supplementing the summarized results in the main text.

\section{Same-Topic and Cross-Topic Editing Results}
\label{appendix:topicresults}

Table~\ref{tab:topics_results} reports detailed editing evaluation results on edited samples, same-topic samples, and cross-topic samples. 
Same-topic samples refer to questions belonging to the same topic category as the edited question, while cross-topic samples correspond to questions from different topic categories. 
The results are reported on a per-user basis.

\section{Significance Test Results}
\label{appendix:significance_test}

We report additional paired Wilcoxon signed-rank test \citep{woolson2007wilcoxon} results for user-level personalization performance across different retrieval settings and evaluation metrics.
Specifically, we evaluate option accuracy, ordinal accuracy, macro-averaged MAE, and CEM under multiple personalization strategies, including model editing, persona augmentation, and retrieval-based few-shot prompting.

All statistical comparisons are conducted against the corresponding retrieval baseline.
Tables~\ref{tab:bm25_user_level_accuracy} and~\ref{tab:contriever_user_level_accuracy} present the detailed significance test results for user-level option accuracy under BM25 and Contriever retrieval settings, respectively.
Tables~\ref{tab:ordinal_accuracy_significance},~\ref{tab:mae_significance}, and~\ref{tab:cem_significance} further report significance test results for ordinal accuracy, macro-averaged MAE, and CEM.

Overall, combining model editing with persona information and retrieval-based few-shot examples consistently achieves the strongest improvements across most evaluation metrics.
The improvements are particularly notable for ordinal accuracy and macro-averaged MAE, where the combined setting demonstrates statistically significant gains over both BM25 and Contriever retrieval baselines.

We additionally compare each retrieval-based personalization baseline with its corresponding model editing variant to evaluate whether model editing consistently improves personalization performance.
Table~\ref{tab:option_accuracy_editing_gain} presents the paired Wilcoxon signed-rank test results for user-level option accuracy between retrieval-based personalization methods and their corresponding editing-enhanced variants.

Across all settings, integrating model editing leads to statistically significant improvements over the corresponding non-editing baselines.
In particular, the largest improvement is observed when applying model editing on top of the Vanilla setting, while retrieval-enhanced and persona-enhanced settings also benefit consistently from model editing.
These results suggest that model editing provides complementary gains beyond retrieval-based personalization alone.

\begin{table*}[t]
\centering
\small

\begin{tabular}{lcccc}
\hline
\textbf{Compared Method} &
\textbf{Baseline Mean} &
\textbf{Method Mean} &
\textbf{Improvement} &
\textbf{Wilcoxon p-value} \\
\hline

Vanilla vs PersonaEdit
& 0.3337 & 0.4467 & +0.1131 & $1.73 \times 10^{-6}$ \\
Few-shot$_{\text{BM25}}$ vs PersonaEdit + Few-shot$_{\text{BM25}}$ & 0.4304 & 0.4575 & +0.0272 & $2.35 \times 10^{-6}$ \\
Persona vs PersonaEdit + persona & 0.4141 & 0.4537 & +0.0396 & $5.14 \times 10^{-6}$ \\
All info$_{\text{BM25}}$ vs PersonaEdit + All info$_{\text{BM25}}$
& 0.4464 & 0.4702 & +0.0238 & 0.00013 \\

\hline
\end{tabular}

\caption{
User-level option accuracy comparison between retrieval-based personalization baselines and their corresponding model editing variants.
All p-values are computed using paired Wilcoxon signed-rank tests.
}
\label{tab:option_accuracy_editing_gain}

\end{table*}

\section{Editing Results with Respect to $k$ and Dataset Size}
\label{appendix:editing_metrics}

Table~\ref{tab:ksize_analysis} and Table~\ref{tab:editing_results_ksize_full} present the detailed editing and locality evaluation results under different hyperparameter settings.

\begin{table*}[t]
\centering
\small
\begin{tabular}{lcccccc}
\toprule
 & \multicolumn{2}{c}{\textbf{20}} & \multicolumn{2}{c}{\textbf{120}} & \multicolumn{2}{c}{\textbf{200}} \\
\cmidrule(lr){2-3} \cmidrule(lr){4-5} \cmidrule(lr){6-7}
Split & Pre-edit & Edited & Pre-edit & Edited & Pre-edit & Edited \\
\midrule
\multicolumn{7}{c}{\textbf{K = 8}} \\
\midrule
Train & 30.67\% & 85.67\% & 34.67\% & 88.75\% & 34.22\% & 89.45\% \\
Test         & 33.37\% & 37.37\% & 33.37\% & 41.75\% & 33.37\% & 44.47\% \\
\midrule
\multicolumn{7}{c}{\textbf{K = 13}} \\
\midrule
Train & 28.83\% & 92.83\% & 33.11\% & 88.50\% & 34.57\% & 89.07\% \\
Test         & 33.37\% & 38.40\% & 33.37\% & 42.67\% & 33.37\% & 44.12\% \\
\midrule
\multicolumn{7}{c}{\textbf{K = 18}} \\
\midrule
Train & 37.00\% & 86.50\% & 31.33\% & 88.61\% & 34.37\% & 88.37\% \\
Test         & 33.37\% & 38.23\% & 33.37\% & 42.63\% & 33.37\% & 44.67\% \\
\bottomrule
\end{tabular}
\caption{Analysis of the clustering hyperparameter K and the scale of edited samples (20, 120, and 200) in terms of accuracy.}
\label{tab:ksize_analysis}
\end{table*} 

\begin{table*}[t]
\centering
\footnotesize
\setlength{\tabcolsep}{2.5pt}
\resizebox{\textwidth}{!}{
\begin{tabular}{llcccccccccc}
\toprule
\multirow{2}{*}{Setting} & \multirow{2}{*}{Method} & \multirow{2}{*}{Efficacy ($\uparrow$)} & \multirow{2}{*}{Paraphrase ($\uparrow$)} & \multirow{2}{*}{Fluency ($\uparrow$)} & \multicolumn{6}{c}{Locality ($\uparrow$)} \\
\cmidrule(lr){6-11}
& & & & & SST & MRPC & CoLA & RTE & NLI & MMLU \\
\midrule
\multirow{2}{*}{$k=8, s=20$} 
& Vanilla   & 40.17$_{{\pm}13.57}$ & 42.33$_{{\pm}12.43}$ & 638.76$_{{\pm}0.00}$ & 95.99$_{{\pm}0.00}$ & 69.81$_{{\pm}0.00}$ & 69.38$_{{\pm}0.00}$ & 22.77$_{{\pm}0.00}$ & 61.62$_{{\pm}0.00}$ & 57.52$_{{\pm}0.00}$ \\
& AlphaEdit & 90.33$_{{\pm}7.52}$  & 62.00$_{{\pm}11.15}$  & 630.02$_{{\pm}2.95}$ & 96.13$_{{\pm}0.43}$ & 65.51$_{{\pm}1.91}$ & 68.83$_{{\pm}1.02}$ & 20.72$_{{\pm}0.66}$ & 62.03$_{{\pm}0.69}$ & 55.57$_{{\pm}1.41}$ \\
\midrule
\multirow{2}{*}{$k=8, s=120$} 
& Vanilla   & 39.56$_{{\pm}6.95}$  & 42.50$_{{\pm}7.39}$  & 640.94$_{{\pm}0.00}$ & 95.99$_{{\pm}0.00}$ & 69.81$_{{\pm}0.00}$ & 69.38$_{{\pm}0.00}$ & 22.77$_{{\pm}0.00}$ & 61.62$_{{\pm}0.00}$ & 57.52$_{{\pm}0.00}$ \\
& AlphaEdit & 89.56$_{{\pm}2.96}$  & 64.25$_{{\pm}5.91}$  & 632.24$_{{\pm}1.08}$ & 96.29$_{{\pm}0.69}$ & 63.54$_{{\pm}1.81}$ & 78.13$_{{\pm}2.35}$ & 20.97$_{{\pm}0.91}$ & 60.38$_{{\pm}1.27}$ & 54.54$_{{\pm}2.06}$ \\
\midrule
\multirow{2}{*}{$k=8, s=200$} 
& Vanilla   & 39.88$_{{\pm}5.94}$  & 40.32$_{{\pm}6.67}$  & 640.98$_{{\pm}0.00}$ & 95.99$_{{\pm}0.00}$ & 69.81$_{{\pm}0.00}$ & 69.38$_{{\pm}0.00}$ & 22.77$_{{\pm}0.00}$ & 61.62$_{{\pm}0.00}$ & 57.52$_{{\pm}0.00}$ \\
& AlphaEdit & 89.18$_{{\pm}1.99}$  & 63.77$_{{\pm}5.14}$  & 631.02$_{{\pm}1.08}$ & 95.69$_{{\pm}0.64}$ & 64.20$_{{\pm}1.66}$ & 78.81$_{{\pm}2.27}$ & 21.44$_{{\pm}1.05}$ & 60.97$_{{\pm}1.46}$ & 55.75$_{{\pm}1.52}$ \\
\midrule
\multirow{2}{*}{$k=13, s=20$} 
& Vanilla   & 40.33$_{{\pm}12.78}$ & 36.17$_{{\pm}12.82}$ & 636.46$_{{\pm}0.00}$ & 95.99$_{{\pm}0.00}$ & 69.81$_{{\pm}0.00}$ & 69.38$_{{\pm}0.00}$ & 22.77$_{{\pm}0.00}$ & 61.62$_{{\pm}0.00}$ & 57.52$_{{\pm}0.00}$ \\
& AlphaEdit & 94.17$_{{\pm}4.67}$  & 62.67$_{{\pm}10.70}$ & 632.68$_{{\pm}2.77}$ & 96.06$_{{\pm}0.25}$ & 65.12$_{{\pm}2.10}$ & 68.96$_{{\pm}1.52}$ & 20.98$_{{\pm}0.73}$ & 62.06$_{{\pm}0.92}$ & 55.73$_{{\pm}1.24}$ \\
\midrule
\multirow{2}{*}{$k=13, s=120$} 
& Vanilla   & 37.56$_{{\pm}6.09}$  & 38.97$_{{\pm}6.07}$  & 642.59$_{{\pm}0.00}$ & 95.99$_{{\pm}0.00}$ & 69.81$_{{\pm}0.00}$ & 69.38$_{{\pm}0.00}$ & 22.77$_{{\pm}0.00}$ & 61.62$_{{\pm}0.00}$ & 57.52$_{{\pm}0.00}$ \\
& AlphaEdit & 88.89$_{{\pm}3.01}$  & 63.39$_{{\pm}5.65}$  & 632.28$_{{\pm}1.38}$ & 96.43$_{{\pm}0.62}$ & 64.11$_{{\pm}1.49}$ & 77.30$_{{\pm}2.80}$ & 21.03$_{{\pm}1.14}$ & 61.64$_{{\pm}1.32}$ & 53.96$_{{\pm}1.66}$ \\
\midrule
\multirow{2}{*}{$k=13, s=200$} 
& Vanilla   & 39.65$_{{\pm}5.77}$  & 41.02$_{{\pm}6.25}$  & 641.15$_{{\pm}0.00}$ & 95.99$_{{\pm}0.00}$ & 69.81$_{{\pm}0.00}$ & 69.38$_{{\pm}0.00}$ & 22.77$_{{\pm}0.00}$ & 61.62$_{{\pm}0.00}$ & 57.52$_{{\pm}0.00}$ \\
& AlphaEdit & 89.47$_{{\pm}2.61}$  & 66.33$_{{\pm}4.74}$  & 630.93$_{{\pm}1.19}$ & 95.79$_{{\pm}0.84}$ & 64.42$_{{\pm}1.30}$ & 78.29$_{{\pm}2.09}$ & 21.77$_{{\pm}0.97}$ & 61.21$_{{\pm}1.66}$ & 55.78$_{{\pm}1.78}$ \\
\midrule
\multirow{2}{*}{$k=18, s=20$} 
& Vanilla   & 46.00$_{{\pm}12.94}$ & 36.50$_{{\pm}11.77}$ & 647.53$_{{\pm}0.00}$ & 95.99$_{{\pm}0.00}$ & 69.81$_{{\pm}0.00}$ & 69.38$_{{\pm}0.00}$ & 22.77$_{{\pm}0.00}$ & 61.62$_{{\pm}0.00}$ & 57.52$_{{\pm}0.00}$ \\
& AlphaEdit & 88.50$_{{\pm}6.21}$  & 67.17$_{{\pm}9.46}$  & 634.54$_{{\pm}2.21}$ & 96.06$_{{\pm}0.36}$ & 65.24$_{{\pm}1.75}$ & 70.77$_{{\pm}1.72}$ & 20.86$_{{\pm}0.83}$ & 61.77$_{{\pm}1.04}$ & 56.39$_{{\pm}1.03}$ \\
\midrule
\multirow{2}{*}{$k=18, s=120$} 
& Vanilla   & 36.78$_{{\pm}6.62}$  & 39.31$_{{\pm}7.08}$  & 641.39$_{{\pm}0.00}$ & 95.99$_{{\pm}0.00}$ & 69.81$_{{\pm}0.00}$ & 69.38$_{{\pm}0.00}$ & 22.77$_{{\pm}0.00}$ & 61.62$_{{\pm}0.00}$ & 57.52$_{{\pm}0.00}$ \\
& AlphaEdit & 89.14$_{{\pm}2.49}$  & 63.22$_{{\pm}6.06}$  & 632.32$_{{\pm}1.18}$ & 96.60$_{{\pm}0.56}$ & 63.16$_{{\pm}1.66}$ & 77.43$_{{\pm}2.85}$ & 21.39$_{{\pm}0.92}$ & 61.43$_{{\pm}1.41}$ & 54.21$_{{\pm}1.87}$ \\
\midrule
\multirow{2}{*}{$k=18, s=200$} 
& Vanilla   & 39.18$_{{\pm}5.64}$  & 39.95$_{{\pm}6.18}$  & 641.11$_{{\pm}0.00}$ & 95.99$_{{\pm}0.00}$ & 69.81$_{{\pm}0.00}$ & 69.38$_{{\pm}0.00}$ & 22.77$_{{\pm}0.00}$ & 61.62$_{{\pm}0.00}$ & 57.52$_{{\pm}0.00}$ \\
& AlphaEdit & 88.72$_{{\pm}2.50}$  & 64.55$_{{\pm}4.72}$  & 630.20$_{{\pm}1.21}$ & 95.56$_{{\pm}0.85}$ & 64.12$_{{\pm}1.83}$ & 78.74$_{{\pm}2.03}$ & 22.07$_{{\pm}0.91}$ & 61.49$_{{\pm}1.86}$ & 56.36$_{{\pm}2.30}$ \\
\bottomrule
\end{tabular}
}
\caption{Editing results of AlphaEdit under various $k$ and $s$ hyperparameter configurations on the LLaMA 3.1 8B Instruct model across efficacy, generalization, fluency, and individual locality tasks.}
\label{tab:editing_results_ksize_full}
\end{table*}

\end{document}